\documentclass{article}
\usepackage{spconf,amsmath,graphicx}

\usepackage {booktabs}
\usepackage{amsmath,amssymb}
\usepackage{bm}
\newcommand{\etal}{\textit{et al}.}
\newcommand{\ie}{\textit{i}.\textit{e}.}

\usepackage{url}

\title{Weakly Semi-supervised Tool Detection \\  in Minimally Invasive Surgery Videos}
\name{Ryo Fujii, Ryo Hachiuma, Hideo Saito}
\address{Keio University, Yokohama, Japan}

\begin{document}
%
\maketitle
\begin{abstract}
Surgical tool detection is essential for analyzing and evaluating minimally invasive surgery videos. Current approaches are mostly based on supervised methods that require large, fully instance-level labels (\ie, bounding boxes). However, large image datasets with instance-level labels are often limited because of the burden of annotation. Thus, surgical tool detection is important when providing image-level labels instead of instance-level labels since image-level annotations are considerably more time-efficient than instance-level annotations. In this work, we propose to strike a balance between the extremely costly annotation burden and detection performance. We further propose a co-occurrence loss, which considers a characteristic that some tool pairs often co-occur together in an image to leverage image-level labels. Encapsulating the knowledge of co-occurrence using the co-occurrence loss helps to overcome the difficulty in classification that originates from the fact that some tools have similar shapes and textures. Extensive experiments conducted on the Endovis2018 dataset in various data settings show the effectiveness of our method. 
\end{abstract}
\section{Introduction}
Surgical tool detection is a fundamental task for recognizing the surgical scene. It can be used for various downstream applications, such as tool tracking, tool pose estimation and skill assessment. To pursue an accurate surgical tool detector, fully-supervised methods with fully-labeled datasets have been investigated~\cite{Jin2018WACV,Zhang2020Access}. However, annotating large-scale object detection datasets is expensive and time-consuming. This may cause the scarcity of labeled surgical tool datasets, and the lack of annotated datasets has essentially hindered the development of accurate surgical tool detection~\cite{Jin2018WACV,Sarikaya2017TMI}.

To reduce data annotation costs, weakly supervised object detection (WSOD) and semi-supervised object detection (SSOD) methods have been proposed in a surgical tool detection task. WSOD methods~\cite{Bilen2016CVPR,Kantorov2016ECCV,Tang2018TPAMI} reduce the cost by replacing the box annotations with large data with cheaper weak annotations, such as image labels, which are much easier to collect than bounding box annotations. While an image label annotation takes one second, a box annotation takes 10 seconds to label an object~\cite{BearmanECCV2016}. On the other hand, SSOD methods~\cite{Jeong2019NEURIPS,Sohn2020Arxiv,Liu2021ICLR,Zhou2021CVPR} train object detectors with a small amount of fully instance-level labeled images and large-scale unlabeled images, which can be collected with significantly lower costs. The core concept behind SSOD is to extract information from unlabeled data. This can be achieved by training a network (\ie~training a teacher model) to solve an object detection task and then leveraging the learned knowledge in a downstream object detection network (\ie~utilizing pseudo-labels generated by a teacher model for training a student model). In surgical tool tasks, only a few works have investigated these approaches ~\cite{Vardazaryan2018MICCAI,Ali2022BBEngIV}. Although both approaches can reduce the annotation cost, their performance is far inferior to their supervised counterparts. It is important to make a trade-off between annotation cost and performance. In this paper, we aim to develop surgical tool detectors with a significantly lower cost of annotation while achieving comparable performance to the fully-supervised approach. We address the task in a weakly semi-supervised manner~\cite{Yan2017Arxiv}, which comprises small fully annotated images and large weakly annotated images by the image-level label.

In the surgical tool detection task, which combines classification and localization (estimating the bounding box position) tasks of surgical tools in surgical videos, the classifier's performance greatly influences the overall object detection performance. While it is easy to localize objects with distinguishable textures from a background, such as a metallic and shiny texture, it is challenging to classify surgical tools that have a similar texture and shape among tools. To address the difficulty in classification, we leverage weak image-level labels. Specifically, we introduce a neural network that refines the category of pseudo-labeled bounding boxes detected from a teacher model. The network is trained with weak image-level labels using Multiple instance learning (MIL). MIL is a weakly supervised learning framework where instance-level ground truths are not observed, but labels for groups of instances (bags) are provided. We conduct MIL regarding instance labels as the category of proposals in a frame and a bag as a weakly annotated image-level label. We adopt the transformer encoder as a component of the refinement model to expect it models the interaction among tools.

\begin{figure}[t] 
\centering
\includegraphics[width=1.0\linewidth]{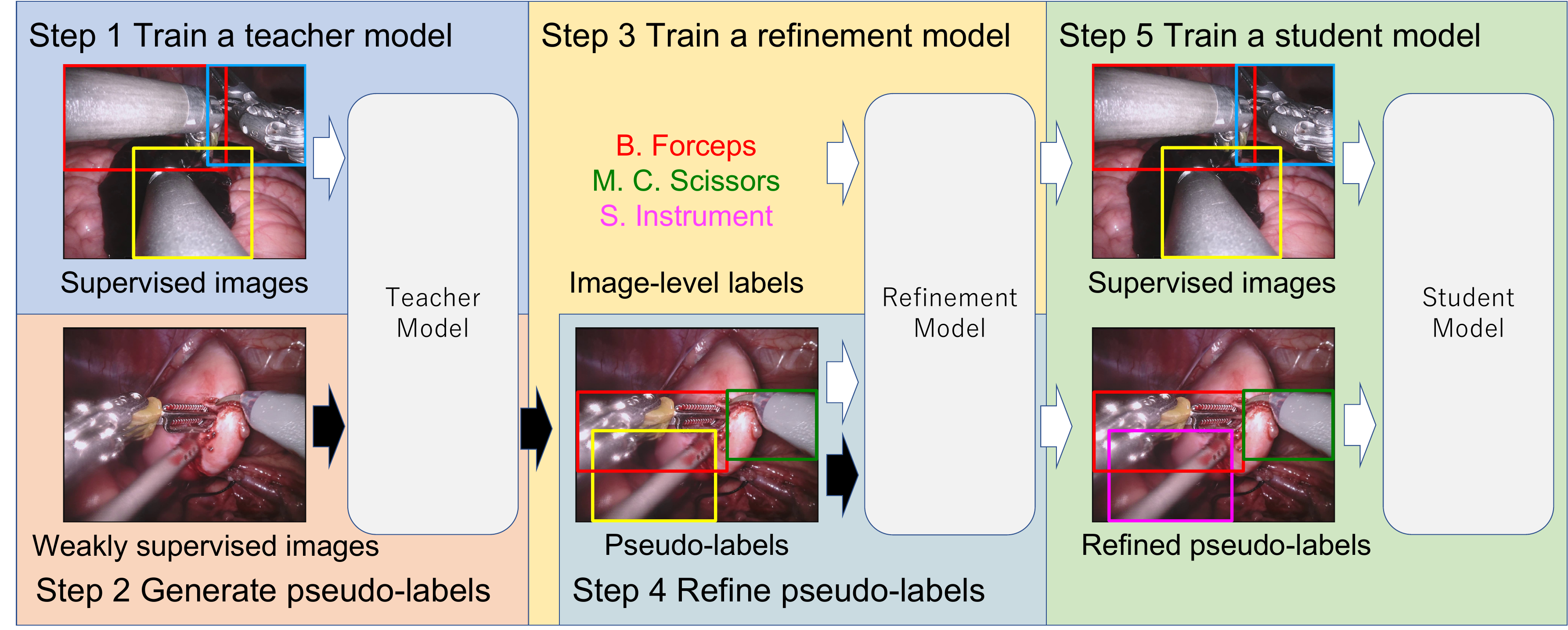}
\caption{Overview of the proposed framework. The white arrows represent the training stage, and the black arrows represent the pseudo-label generation stage.}\label{fig:overview}
\end{figure}

Furthermore, we introduce a co-occurrence loss to enhance the performance of the refinement model, leveraging the observation that certain surgical tool pairs often co-occur. For instance, in the Endovis2018 dataset, pairs of bipolar forceps and monocular curved scissors frequently appear together in an image, used for coagulating tissues to stop bleeding and for dissection, respectively. Incorporating this relational context, which represents the statistical tendency of co-occurrence in images, enables the network to learn relationships among tools. This information can serve as a valuable prior for the tool classification task.

We employ the Endovis2018 dataset~\cite{Allan2020Robot} to demonstrate the effectiveness of the proposed learning framework. We compare our method with a semi-supervised baseline in a setting where object instances of small image data fractions are fully annotated, and image-level labels annotate the rest. Our proposed detector outperforms the detector trained with the baseline method with a different fraction of fully-annotated image data. In particular, when using $27\%$ fully labeled data, our learning framework enhances the mean average precision by $10.7$ percentage points. The performance of the detector is comparable to the detector trained in a fully supervised manner. Finally, the ablation study shows the effectiveness of the proposed co-occurrence loss.

 \begin{figure}[tb]
\begin{center}
\resizebox{\columnwidth}{!}{
\begin{tabular}{c}
\begin{minipage}{1.0\hsize}
    \begin{center}
        \includegraphics[trim={0 1em 0 0}, clip, width=\hsize]{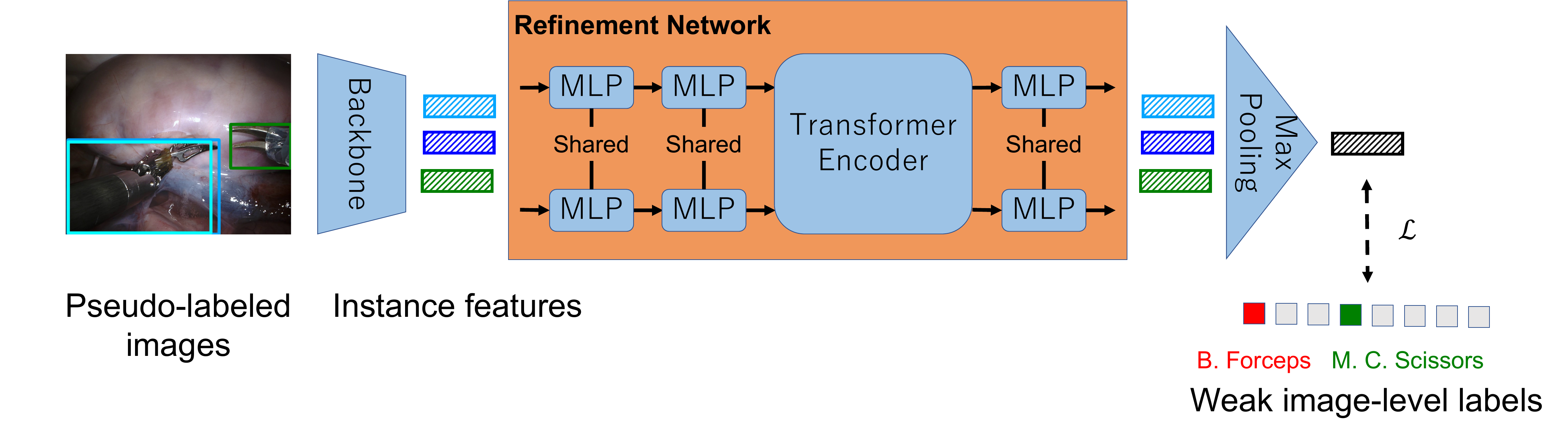} 
        {\footnotesize (a) Training a refinement model in the MIL manner.}
    \end{center}
\end{minipage}
\\
\begin{minipage}{1.0\hsize}
    \begin{center}
        \includegraphics[trim={0 1em 0 0}, clip, width=\hsize]{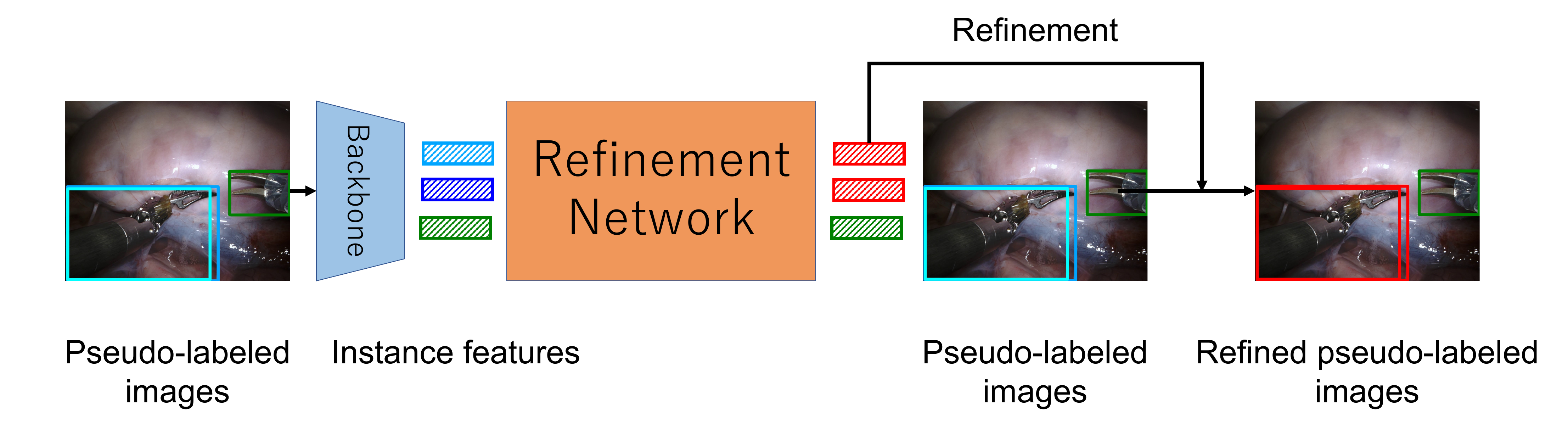}
        {\footnotesize (b) Refinment of pseudo-labels.}
    \end{center}
\end{minipage}
\end{tabular}}
\caption{Details of training and refinement procedure of a refinement model corresponding to step 3 and step 4 in our framework.}
\label{fig:refmodelarch}
\end{center}
\end{figure}

\section{Approach}
\subsection{Problem Definition} 
In this work, we study weakly semi-supervised object detection under an image-level annotated setting, where the training dataset consists of a small set of fully annotated images denoted as $D_f = \{I_i, {\bm b}_i\}_{i=1}^{N_f}$ and a large set of
image-level annotated images denoted as $D_c = \{I_i, {\bm y}_i\}_{i=1}^{N_c}$. $N_f$ and $N_c$ are a number of fully annotated and wealy annotated samples, respectively. ${\bm b}_i$ represents 
a set of bounding box annotations ${\bm b}_i = \{u_1^j, v_1^j, u_2^j, v_2^j, c^j\}_{j=1}^{J}$ (tool position $u_1, v_1, u_2, v_2$ and the category $c$) of each labeled image $I_i$, and ${\bm y}_i$ denotes the weakly-labeled annotation for each image.

\subsection{Overall Framework} 
The overall training pipeline is divided into the following five stages. (1) train a teacher model on a fully-annotated dataset $D_f$, (2) generate pseudo-labels (bounding box and the category) using the trained teacher model on a weakly-annotated dataset $D_c$, (3) train a refinement model, which refines the category of pseudo-labels in a MIL manner using the weakly annotated image-level labels, (4) generate refined pseudo-labels using the trained refinement model, and (5) train a student model with fully labeled images and pseudo-labeled images. The overall framework is shown in Figure \ref{fig:overview}. We employ the Faster R-CNN~\cite{Ren2015NIPS} with Feature Pyramid Networks (FPN)~\cite{Lin2017CVPR} for both the teacher model and the student model.

\subsection{Refinement Network} 
The refinement model aims to refine the category label of the pseudo-labels so that the student network learns the correctly refined category labels. In stage 2, the bounding boxes and the corresponding visual features $x_i$ are extracted from the intermediate RoI Align layer~\cite{He2017ICCV} in Faster R-CNN from input image $I_i$. These visual features are forwarded to two MLP encoders with shared weights and five transformer encoders~\cite{Vaswani2017NIPS}, which perform tool interaction reasoning using self-attention. The last MLP layer predicts the category of each proposal. The architecture is shown in Figure \ref{fig:refmodelarch} (a).
 
\subsection{Training Refinement Model in Multiple Instance Learning} 
In the MIL setting, one is given a bag of $N$ instances, denoted as $x={x_1, x_2, \ldots, x_N}$. A bag is defined as positive if it contains at least one positive instance (however, it is not known which one is positive) and negative otherwise. For the multi-label classification problem, the label vector for the bag is $y \in \mathbb{R}^c$, and $y_k = 1$ if there is at least one instance with the $k$th label present in the bag, and $y_k = 0$ otherwise.

Our goal is to train an instance-level classifier that predicts the label probabilities for the $j$th instance, which is represented by a probabilistic form: $p(y_k|x_j)$. We use the aggregation function $g(\cdot)$ to aggregate the set of instance-level probabilities to predict the bag-level probabilities:
\begin{equation}
p(y_{k}=1|x_1, x_2,...,x_N) = g(p_{1}, p_{2},...,p_{N}),
\end{equation}
where $p(y_{k}=1)$ represents the bag-level probability for the $k$th label, and $p_{i}$ represents the instance-level label probability of the $j$th instance. We choose max-pooling as an aggregation function:
\begin{equation}
g({p_i}) = \max_j p_j.
\end{equation}
Max-pooling considers only the top-scoring instance in the bag, which effectively accounts for the assumption that at least one instance in the bag has the specified bag-level label.

After the aggregation procedure, we can obtain the bag-level predictions and can apply a standard multi-label classification loss. We use the binary cross-entropy loss:
\begin{equation}
\mathcal{L}_{ce} = -\sum_{k}^{C} y_{k}\log p_{k} + (1-y_{k}) \log (1-p_{k}),
\end{equation}
where $p_{k}$ represents the bag-level probability for the $k$th label, and $C$ is the number of classes.

\subsection{Co-occurence Loss} 
Bengio \etal~\cite{bengio2013arxiv} propose the Ising-like penalty to incorporate co-occurrence statistics from web documents into the model to improve classification and detection accuracy. Inspired by this work, we present the co-occurrence loss, which makes the model consider the statistical tendency of co-occurrence in minimally invasive surgery videos:
\begin{equation}
\mathcal{L}_{co} = - \sum_{k}^{C} p_{k}^\mathrm{T}Sp_{k},
\end{equation}
where $p_{k}$ represents the bag-level probability for the $k$th label, and $C$ is the number of classes. Each element of $S_{i,j}$ is constructed by the point-wise mutual information:
\begin{equation}
s_{i,j} = \log \frac{p(i,j)}{p(i)p(j)},
\end{equation}
where $p(i)$ and $p(i,j)$ represent the probability of occurrence of class $i$ and the probability of co-occurrence of class $i$ and $j$, respectively.
We then transform the scores using the logit function:
\begin{equation}
  S_{i,j}=
  \begin{cases}
    \frac{1}{1+\exp(s_{i,j})} & \text{if $s_{i,j}>0$,} \\
    0                 & \text{if otherwise.,} 
  \end{cases}
\end{equation}

The co-occurrence loss captures the co-occurrence of object category pairs. When $S_{i,j}$ is high, the two categories tend to co-occur in a frame. We compute the $S_{ij}$ using the weak image-level labels.

\subsection{Loss Function} 
We combine binary cross-entropy and co-occurrence loss to train the refinement model as follows;
\begin{equation}
\mathcal{L}= \mathcal{L}_{ce} + \alpha \mathcal{L}_{co},
\end{equation}
where $\alpha$ is set to $0.0001$ in our experiments.

\subsection{Category Label Refinement}
After training the refinement model, we can obtain the instance-level classifier. We apply it to the instances proposed by the teacher model in the form of pseudo-labels, then we adopt the refined class label predicted from the refinement model as the category of instances, as shown in Figure \ref{fig:refmodelarch} (b). Finally, we utilize the refined pseudo-labels for the training of the student model.

\section{Experiments}
We evaluate our models on the Endovis2018 dataset~\cite{Allan2020Robot}. We report the standard object detection metrics, including mAP (averaged over different IoU thresholds), followed by the evaluation metrics in MS-COCO~\cite{Lin2014ECCV}.
\subsection{Dataset}
We benchmark our proposed model on the EndoVis2018 dataset~\cite{Allan2020Robot} released at the Robotic Scene Segmentation Challenge. The dataset consists of $15$ sequences, each composed of $149$ frames. The input image size is $1280\times1024$. We use the annotated masks of the tool type provided by Gonz{'a}lez \etal~\cite{Gonzlez2020MICCAI} and follow the procedures taken by Sanchez \etal~\cite{Sanchez2021MICCAI} to generate bounding boxes. Following ISINet, we divide the dataset into two sets, the validation set with sequences $2$, $5$, $9$, and $15$, and the training set with the remaining ones. We randomly sample the sequences from the training set as the fully labeled set and use the rest as a weakly labeled set. Hence, we evaluate our model with seven settings, where the $27\%$, $36\%$, $45\%$, $54\%$, $63\%$, $72\%$, and $81\%$ in the training set are used as the fully-labeled dataset.

\subsection{Implementation Details}
In our framework, there are three models: the teacher model, the refinement model, and the student model. We adopt the Faster R-CNN model with FPN~\cite{Lin2017CVPR} for both the teacher and student models. We utilized the implementations of Detectron2~\cite{Wu2019GitHub}. For the training of both teacher and student models, we fine-tune models pre-trained on MS-COCO~\cite{Lin2014ECCV} with a batch size of $16$. We set the learning rate to $0.01$ and weight decay to $0.0001$. The networks were trained for $3$K iterations. We report the results on the validation set obtained in the last epoch. These settings remain fixed for all the experiments.

We employ the SGD optimizer to train a refinement model with a learning rate of $5.0\times10^{-3}$ with cosine annealing decay. We utilize the pre-trained feature extractor and MLP layers from a teacher model. During training, we freeze the feature extractor weights. The refinement model is trained for 50 epochs. Note that the size of the batch size varies for every image, as the number of detected surgical tools determines the batch size. All training is conducted on a single NVIDIA RTX A5000 GPU.

\begin{figure}[tb]
\centering
\begin{tabular}{cc}
\begin{minipage}{0.48\hsize}
    \begin{center}
        \includegraphics[clip, width=\hsize]{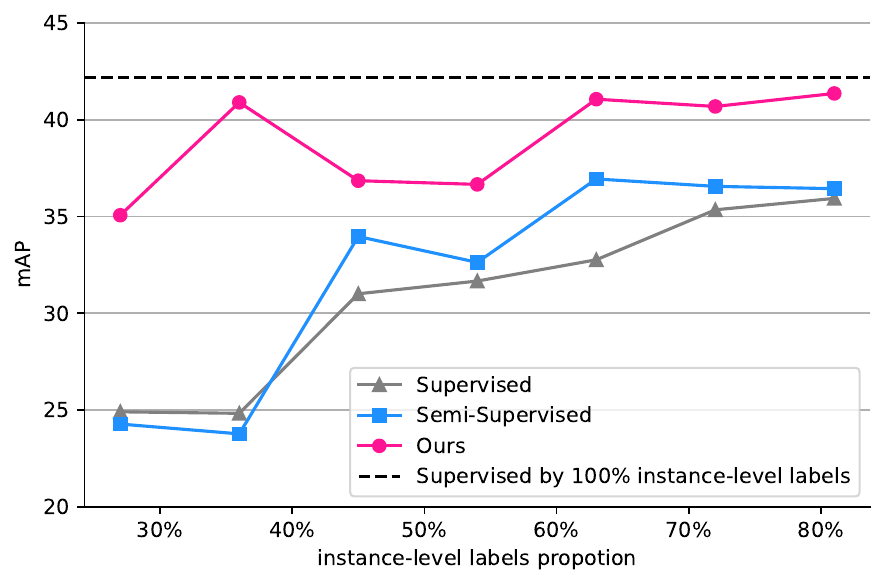} 
    \end{center}
\end{minipage}
&
\begin{minipage}{0.48\hsize}
    \begin{center}
       \includegraphics[clip, width=\hsize]{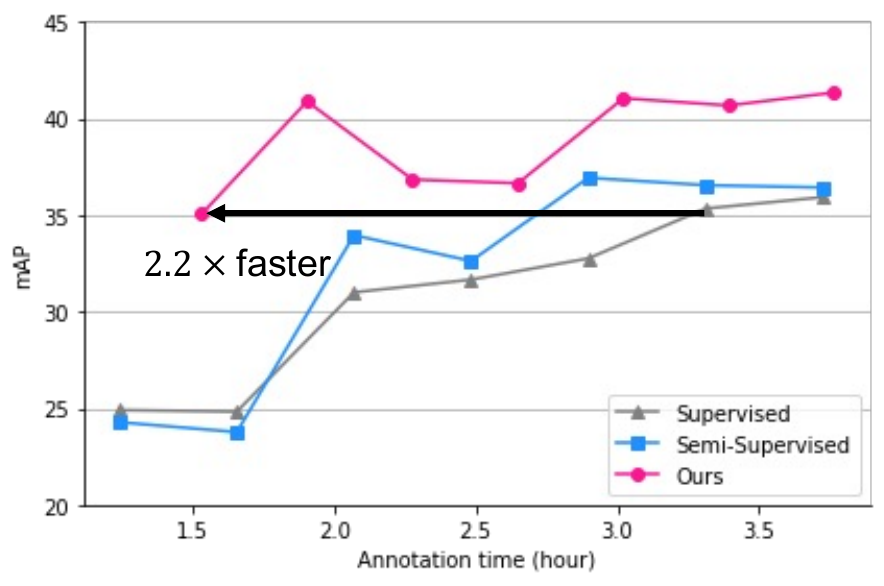}
    \end{center}
\end{minipage}
\end{tabular}
\caption{Comparison in mAPs of the student model (\ie~Faster-RCNN) for different supervision on Endvis2018. 'Supervised' and 'Semi-Supervised' refer to the student models trained on labeled data only and trained on labeled data and pseudo-labels obtained from a teacher model without refinement, respectively.}
\label{fig:comparison}
\end{figure}

\subsection{Comparison with the Baselines}
\begin{figure}[tb]
\centering
\scalebox{0.86}{
\begin{tabular}{cccc}
\begin{minipage}{0.15\hsize}
    \begin{center}
       Ground Truth
    \end{center}
\end{minipage}
&
\begin{minipage}{0.15\hsize}
    \begin{center}
         Supervised
    \end{center}
\end{minipage}
&
\begin{minipage}{0.15\hsize}
    \begin{center}
        Semi-supervised
    \end{center}
\end{minipage}
&
\begin{minipage}{0.15\hsize}
    \begin{center}
        Ours
    \end{center}
\end{minipage}
\\
\begin{minipage}{0.25\hsize}
    \begin{center}
        \includegraphics[clip, width=\hsize]{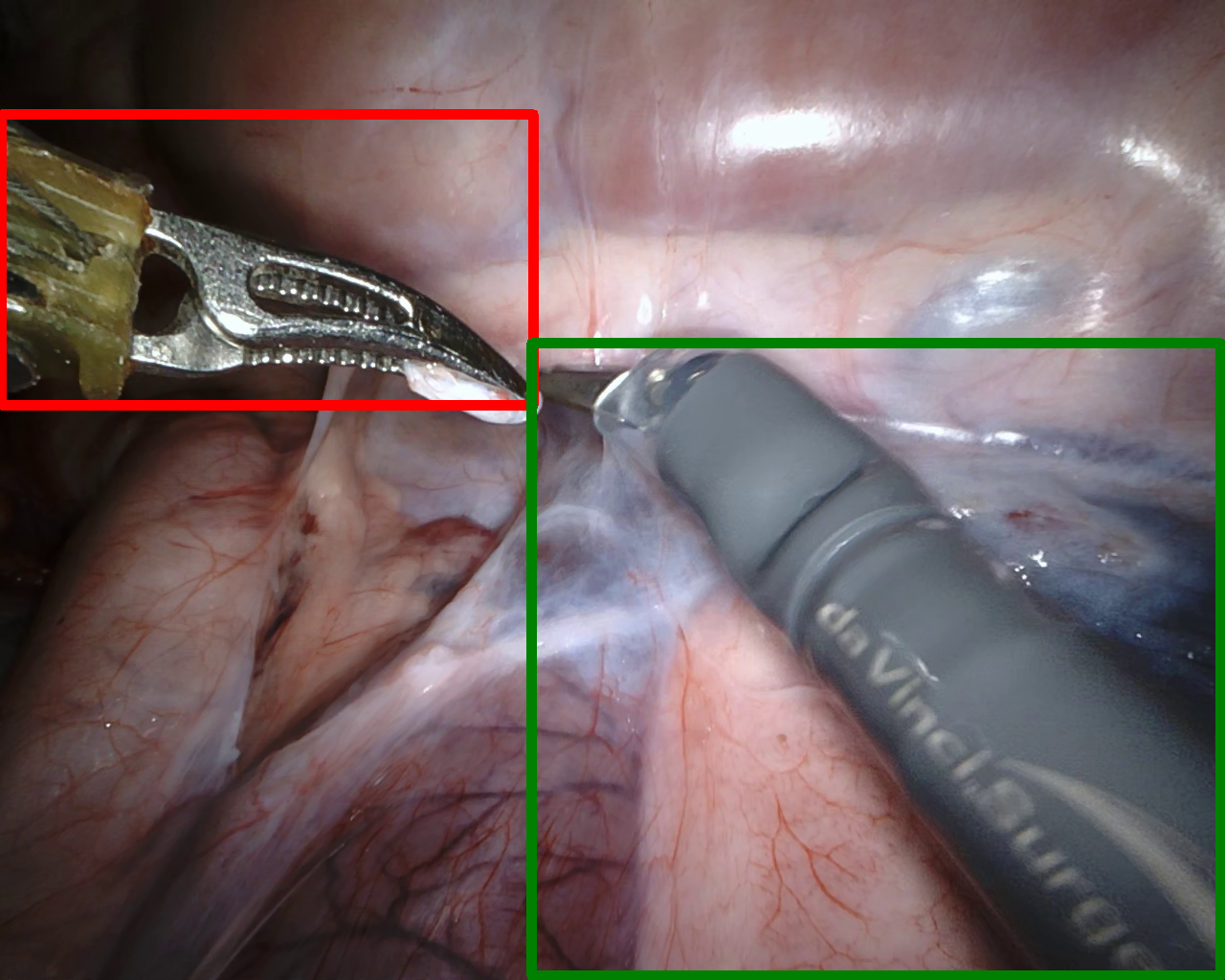} 
    \end{center}
\end{minipage}
&
\begin{minipage}{0.25\hsize}
    \begin{center}
       \includegraphics[clip, width=\hsize]{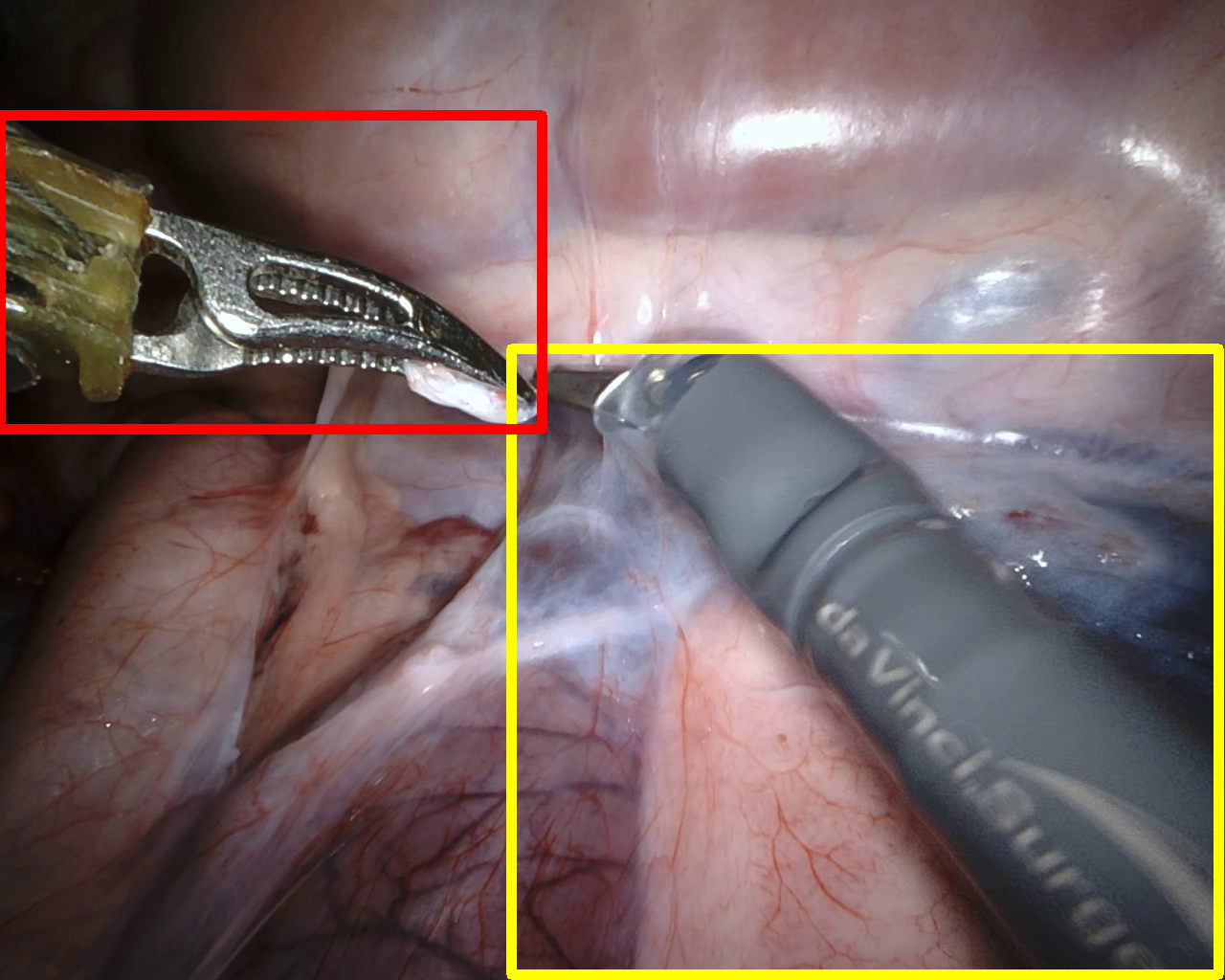} 
    \end{center}
\end{minipage}
&
\begin{minipage}{0.25\hsize}
    \begin{center}
       \includegraphics[clip, width=\hsize]{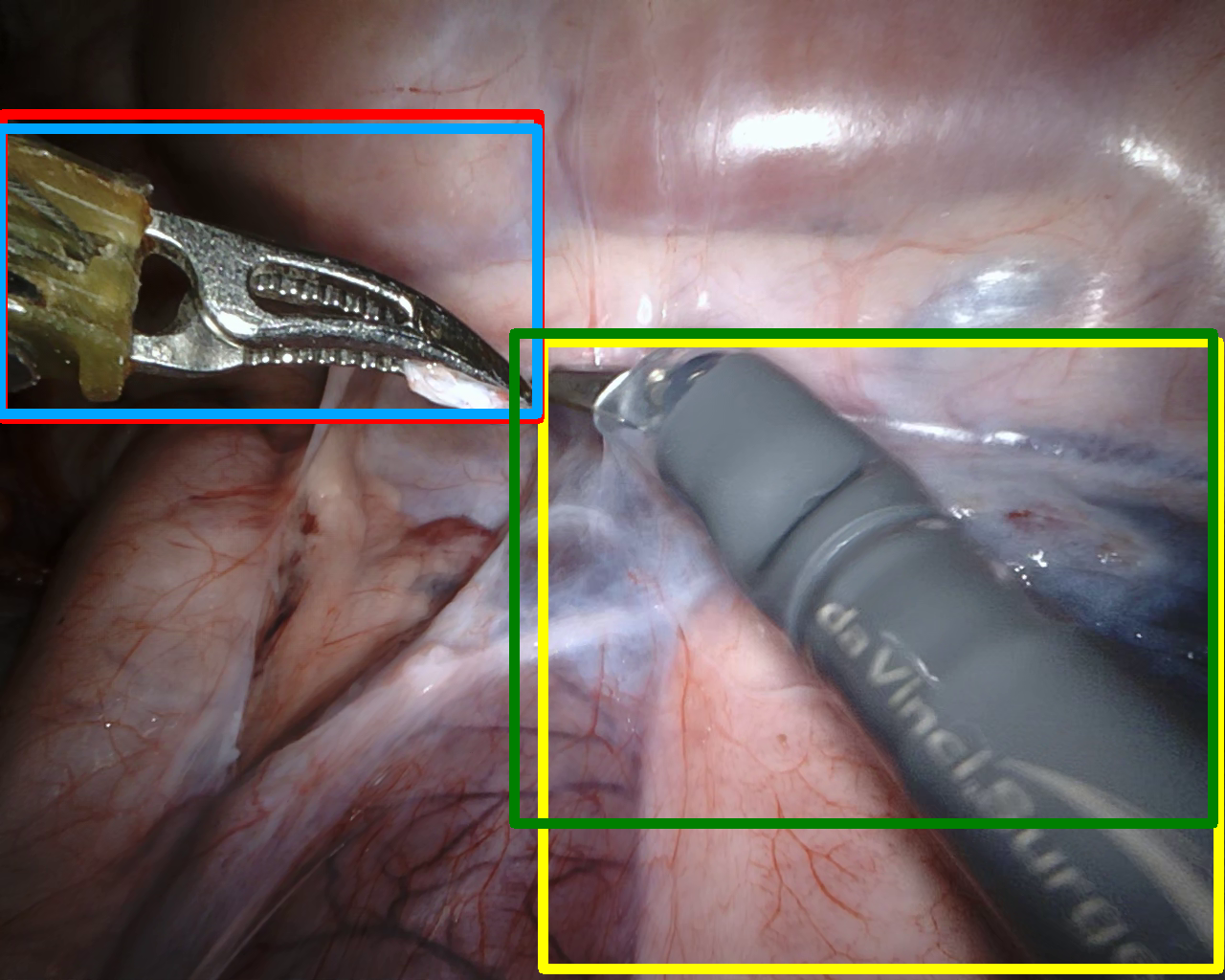} 
    \end{center}
\end{minipage}
&
\begin{minipage}{0.25\hsize}
    \begin{center}
        \includegraphics[clip, width=\hsize]{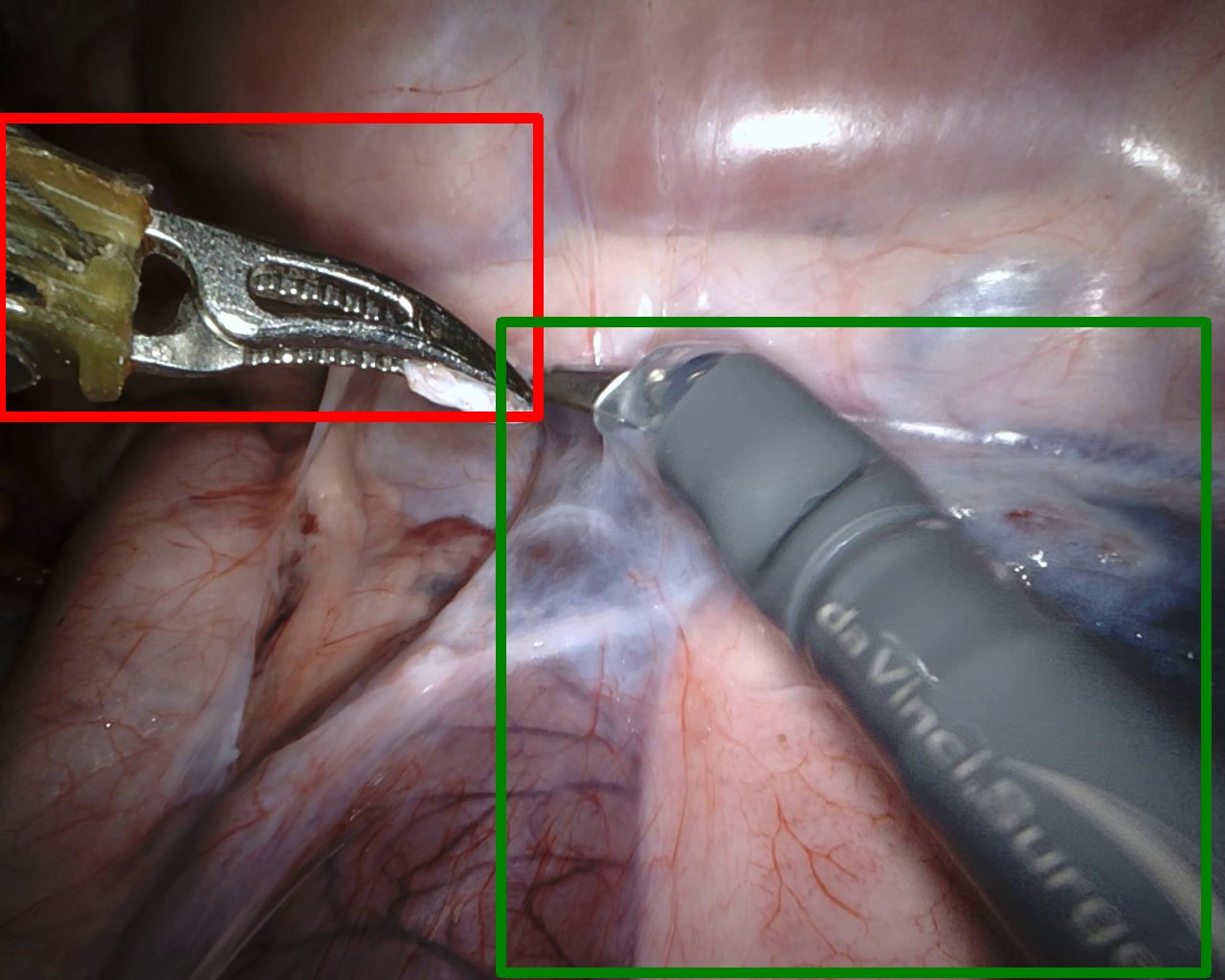} 
    \end{center}
\end{minipage}
\\
\begin{minipage}{0.25\hsize}
    \begin{center}
     \includegraphics[clip, width=\hsize]{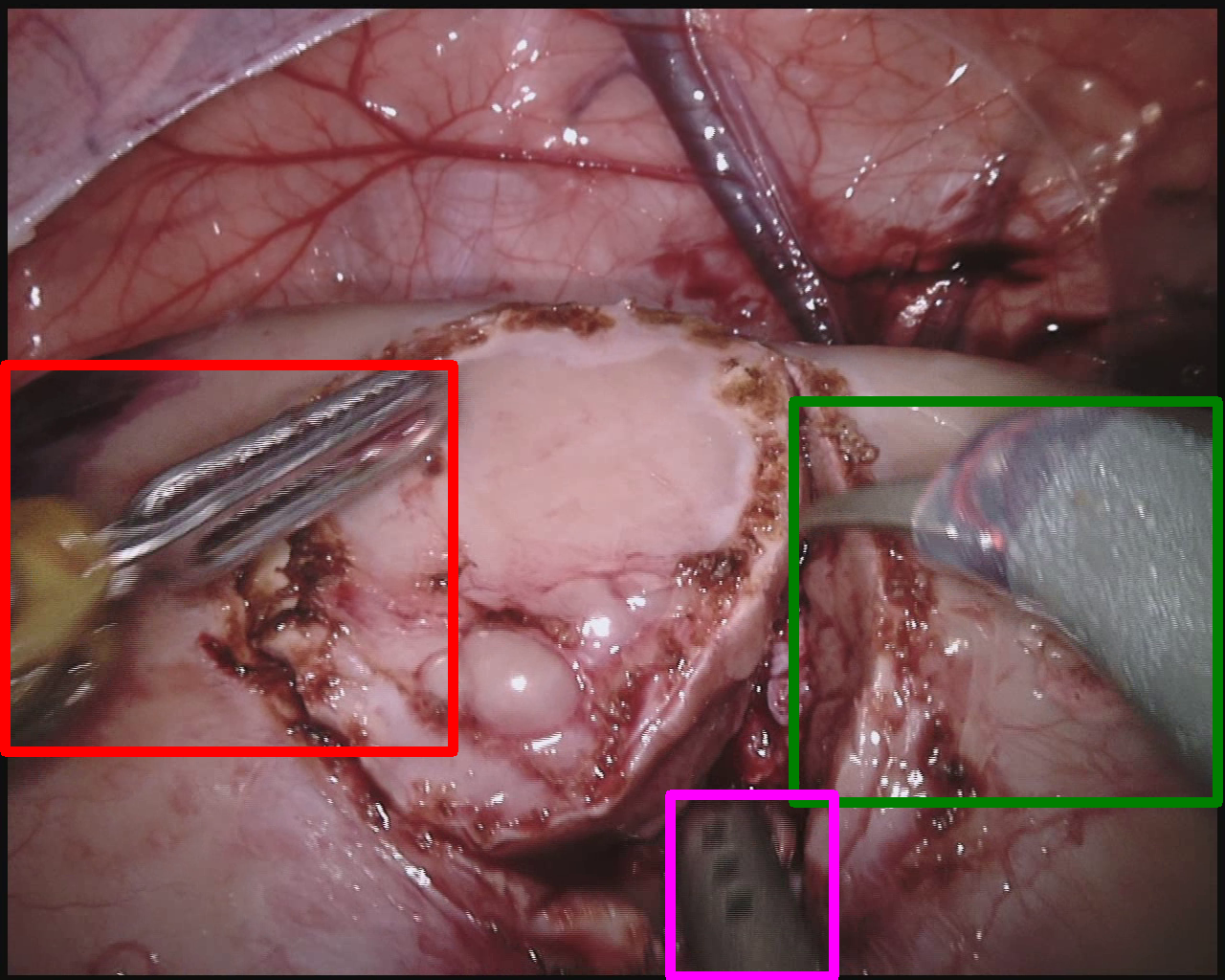} 
    \end{center}
\end{minipage}
&
\begin{minipage}{0.25\hsize}
    \begin{center}
        \includegraphics[clip, width=\hsize]{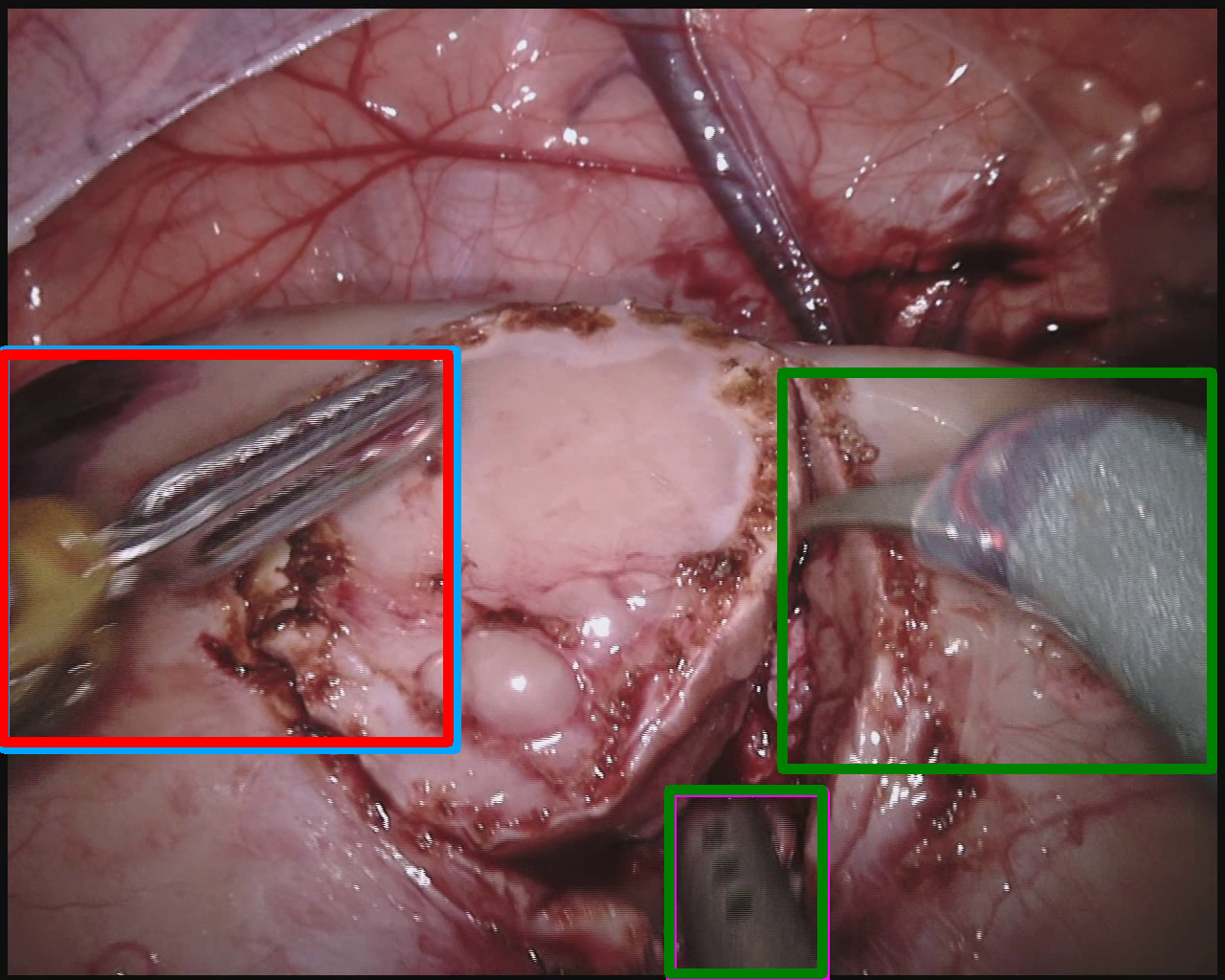} 
    \end{center}
\end{minipage}
&
\begin{minipage}{0.25\hsize}
    \begin{center}
        \includegraphics[clip, width=\hsize]{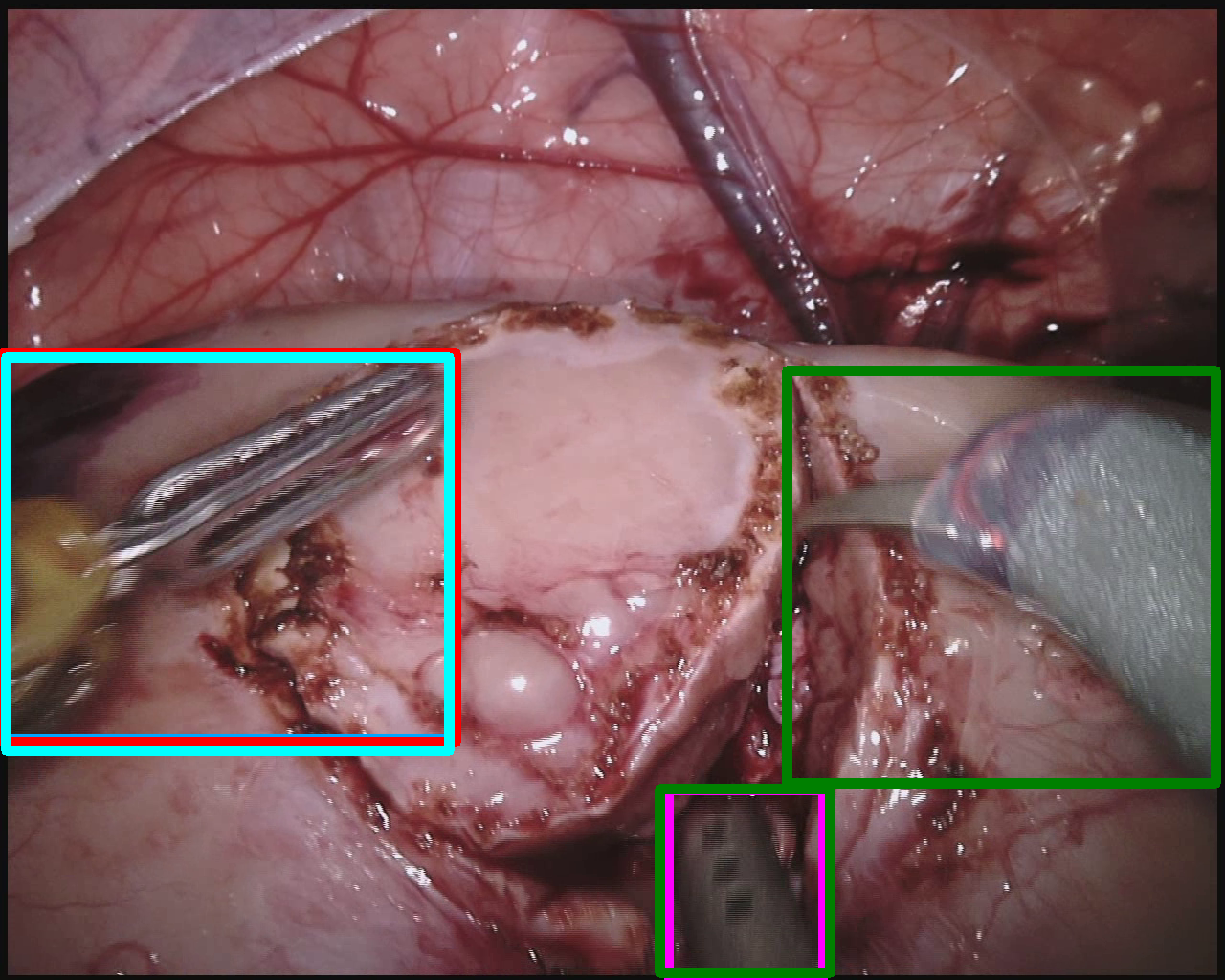} 
    \end{center}
\end{minipage}
&
\begin{minipage}{0.25\hsize}
    \begin{center}
        \includegraphics[clip, width=\hsize]{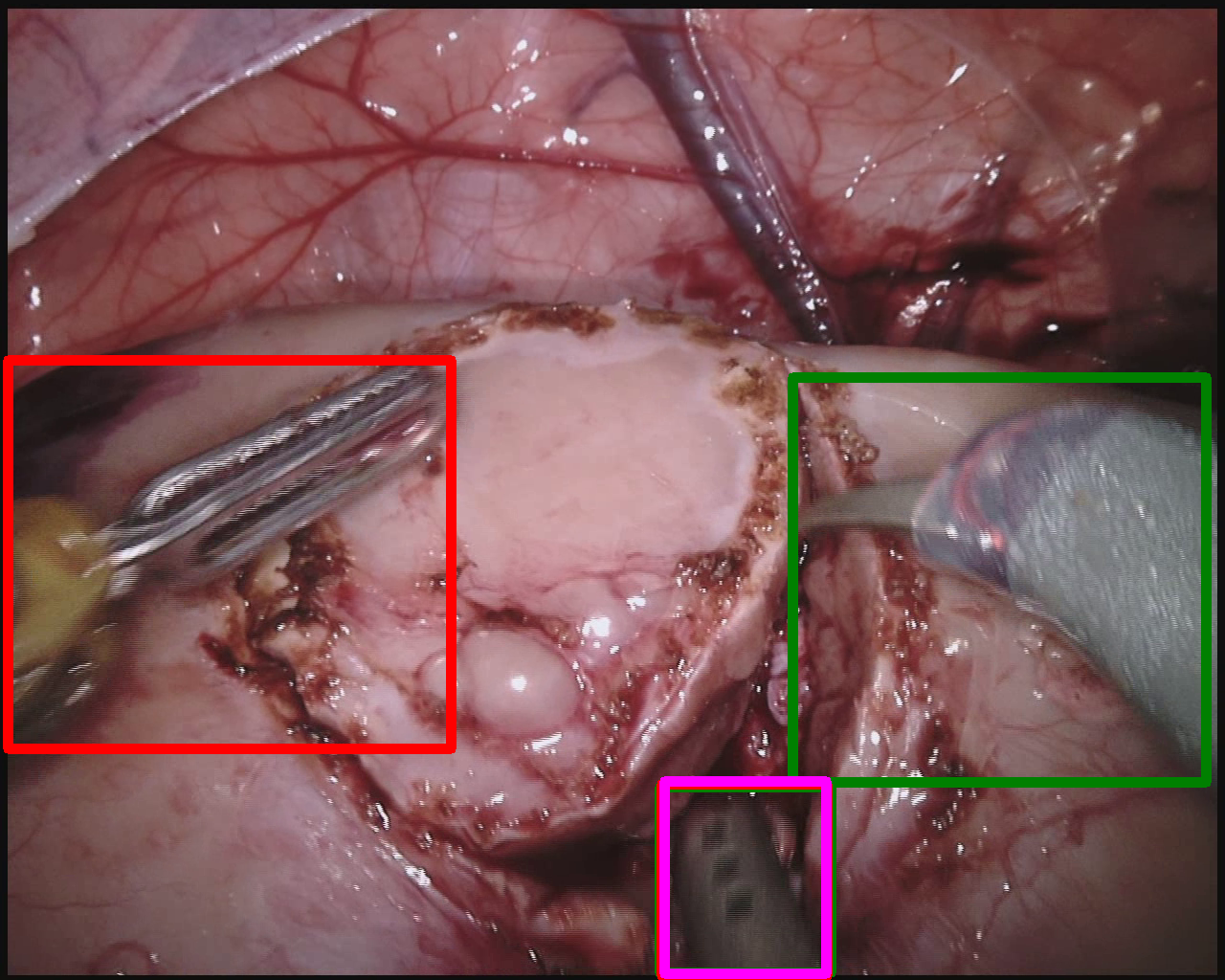} 
    \end{center}
\end{minipage}
\\
\begin{minipage}{0.25\hsize}
    \begin{center}
     \includegraphics[clip, width=\hsize]{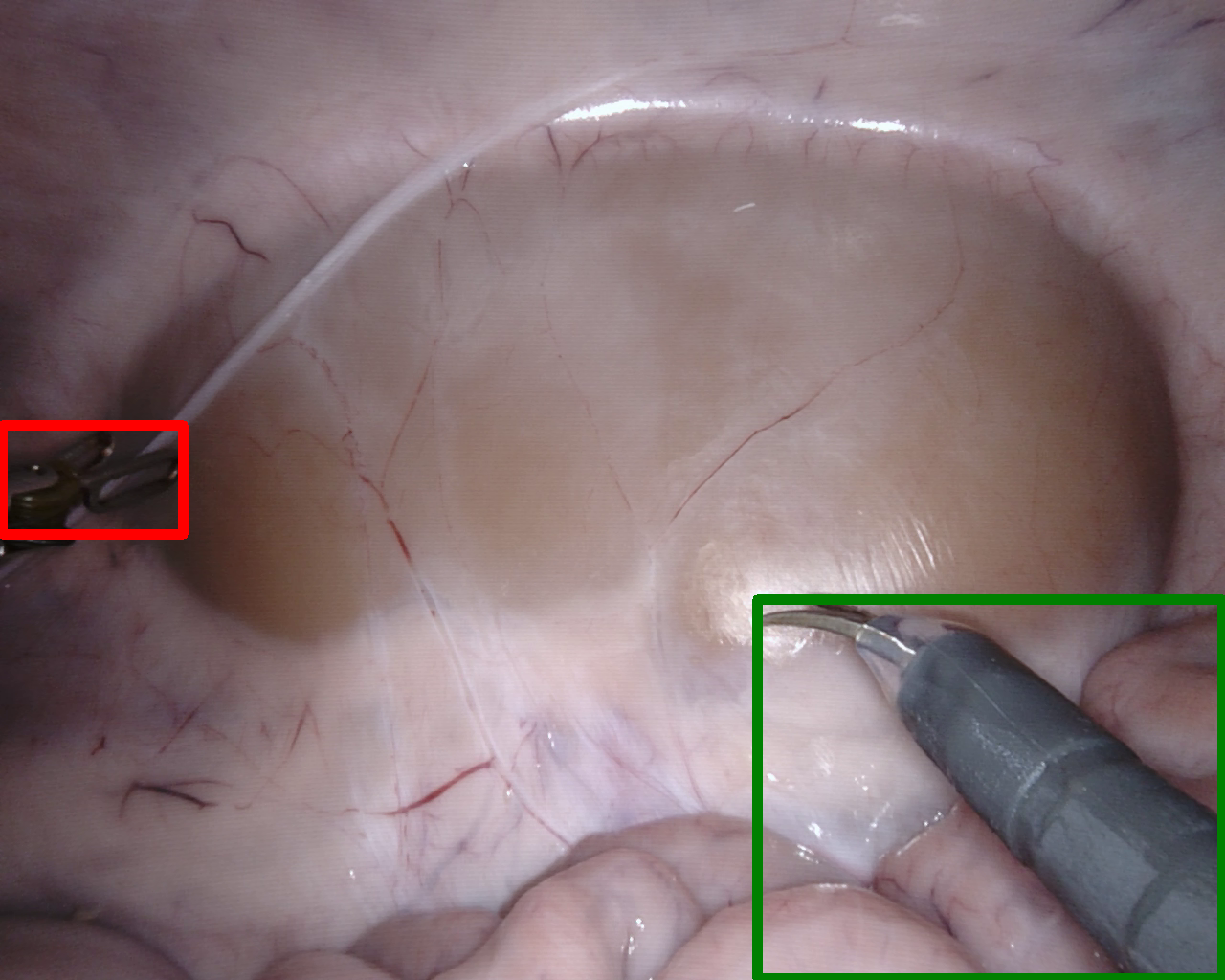} 
    \end{center}
\end{minipage}
&
\begin{minipage}{0.25\hsize}
    \begin{center}
        \includegraphics[clip, width=\hsize]{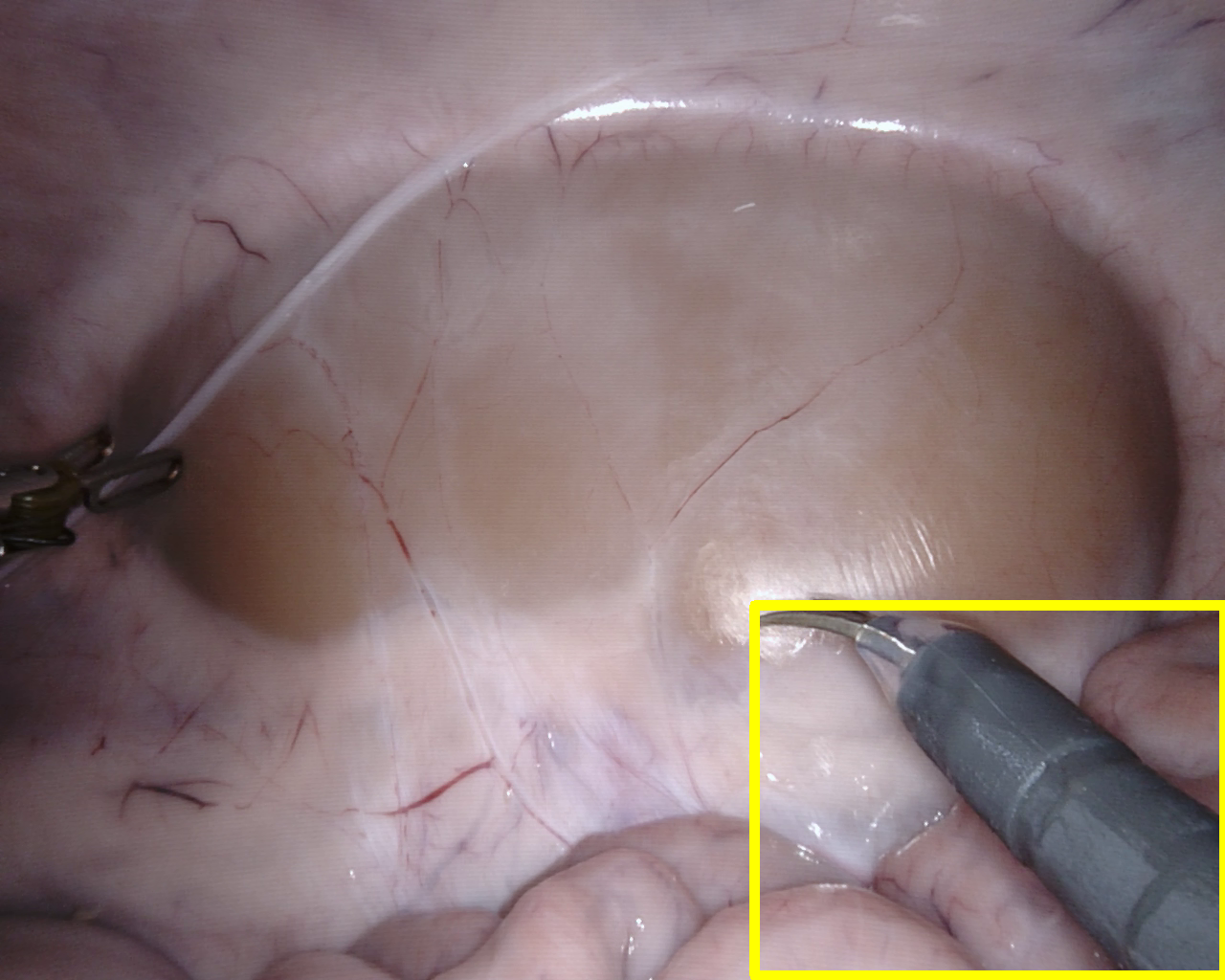} 
    \end{center}
\end{minipage}
&
\begin{minipage}{0.25\hsize}
    \begin{center}
        \includegraphics[clip, width=\hsize]{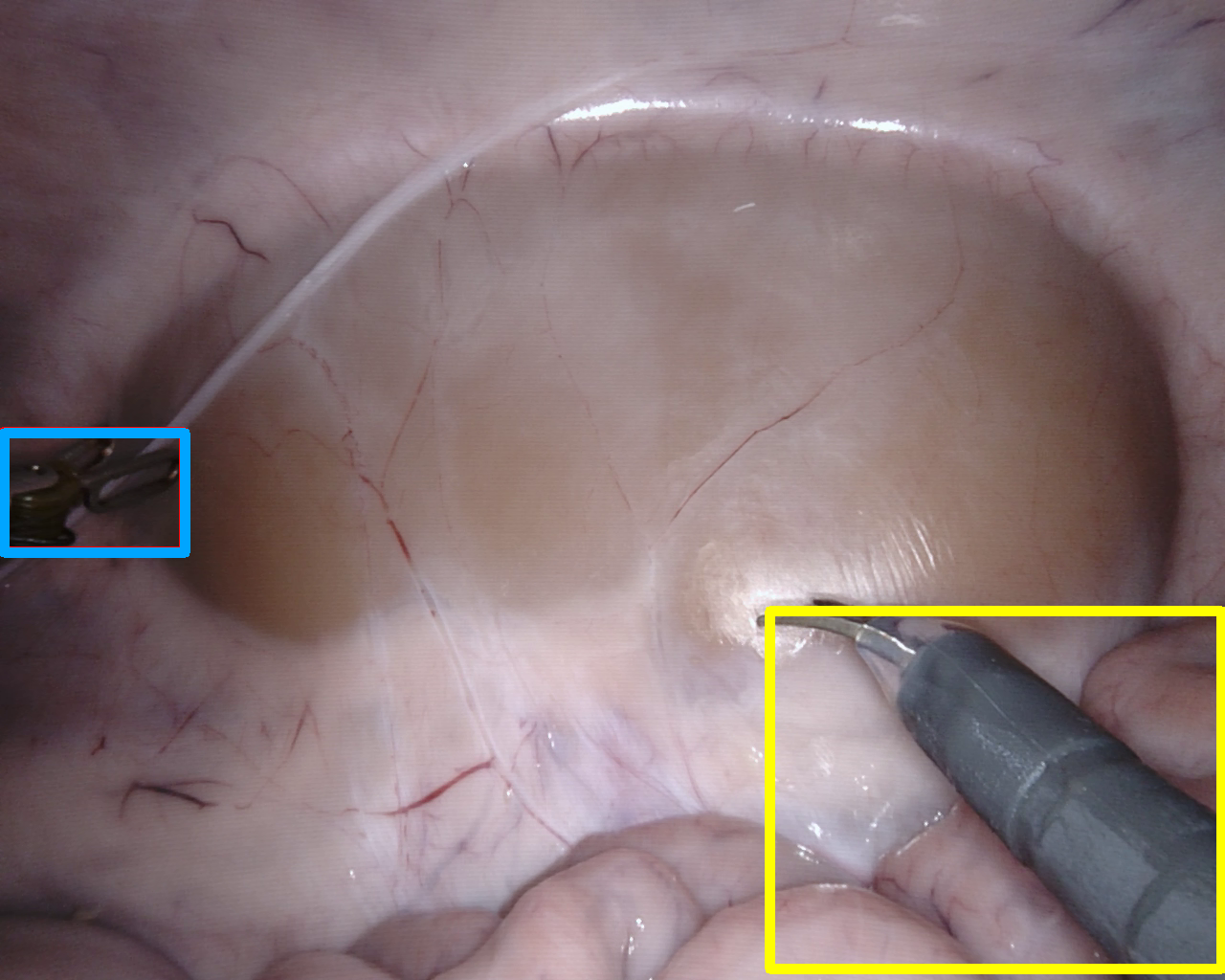} 
    \end{center}
\end{minipage}
&
\begin{minipage}{0.25\hsize}
    \begin{center}
        \includegraphics[clip, width=\hsize]{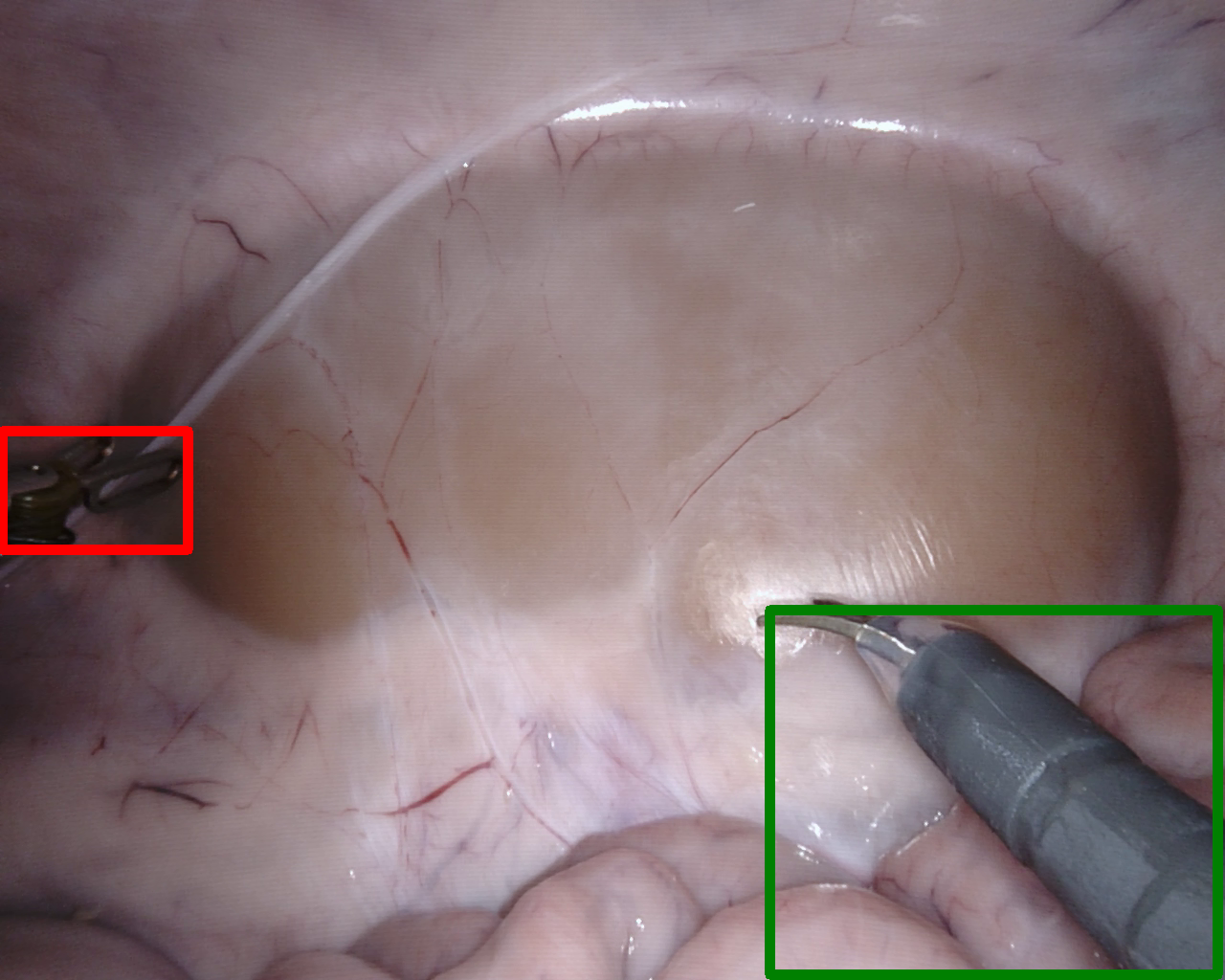} 
    \end{center}
\end{minipage}
\end{tabular}
}
\caption{Qualitative results of the object detection with different supervision methods. The colors of the bounding boxes denote the estimated category of the tools.}
\label{fig:vis}
\end{figure}

We evaluate our weakly semi-supervised framework against different supervision methods. As baselines, we train the student model (\ie~Faster-RCNN) only with the fully annotated images (denoted as 'Supervised') and with both the fully annotated images and pseudo-labels without refinement (denoted as 'Semi-Supervised'). Note that among the baselines and the proposed method, the models are trained with different training schemes, and the network architecture and the number of parameters (Faster R-CNN) at the test time are identical. Figure~\ref{fig:vis} summarizes the qualitative results of the proposed framework and comparison with the baselines, and our method accurately classifies the instance.

Figure~\ref{fig:comparison} (left) summarizes the results of the methods in different data split settings. The proposed and semi-supervised frameworks that utilize the pseudo-labels outperform the supervised framework, which shows the benefits of the pseudo-labels. Our method outperforms the semi-supervised framework by a considerable margin ($25.0\%$ vs. $35.1\%$ mAP when $30\%$ fully-labeled data). This demonstrates that pseudo-label refinement using image-level labels can improve the performance of the student model.

Moreover, the x-axis of Figure~\ref{fig:comparison} (right) shows the time to prepare the labeled dataset. We calculate the annotation time of bounding boxes and the image label based on the conventional work~\cite{BearmanECCV2016}. It shows that our framework with image-level labels significantly outperforms models trained with the other supervision forms under the same annotation budget.

\subsection{Ablation Study}
As an ablation study, we explore the impact of our proposed co-occurrence loss on the model performance in different data splits. As shown in Table \ref{fig:ab}, when using the co-occurrence loss, the mAP of the student model is improved by $1.4$ percentage points on average

\begin{table}[tb]
\caption{Comparison of the effectiveness of the co-occurrence loss on different labeled proportions. Results are reported with mAP.}
\label{table:5}
\begin{center}
\resizebox{\columnwidth}{!}{
\begin{tabular}{c||ccccccc|c}
\toprule
Proportion & 27\% & 36\%  & 45\%  & 54\% & 63\%  & 72\% & 81\% & Ave.\\
 \hline
Only BCE loss & \textbf{38.1} & 32.4 & \textbf{36.3} & 36.3 & 40.2 & 39.3 & 39.8 & 37.5 \\
 BCE and co. loss (Ours) & 35.1 & \textbf{40.9} & 36.8 & \textbf{36.7} & \textbf{41.0} & \textbf{40.7} & \textbf{41.4} & \textbf{38.9}\\
\bottomrule
\end{tabular}}
\label{fig:ab}
\end{center}
\end{table}

\section{Conclusion}
In this work, we address the task of tool detection in the weakly semi-supervised learning task. We introduce a neural network that refines the category of pseudo-labeled bounding boxes detected from a teacher model. The network is trained with weak image-level labels using MIL. We show the effectiveness of refinement for the performance of a student model. We also present the co-occurrence loss, which incorporates relation context among tools. Our approach is simple and can be easily introduced to existing detectors. We demonstrate its efficacy through extensive experimental analysis.

\bibliographystyle{IEEEbib}
\bibliography{refs}

\begin{thebibliography}{10}

\bibitem{Jin2018WACV}
Amy Jin, Serena Yeung, Jeffrey Jopling, Jonathan Krause, Dan Azagury, Arnold Milstein, and Li~Fei-Fei,
\newblock ``{Tool Detection and Operative Skill Assessment in Surgical Videos Using Region-Based Convolutional Neural Networks},''
\newblock in {\em WACV}, 2018.

\bibitem{Zhang2020Access}
Beibei Zhang, Shengsheng Wang, Liyan Dong, and Peng Chen,
\newblock ``{Surgical Tools Detection Based on Modulated Anchoring Network in Laparoscopic Videos},''
\newblock {\em IEEE Access}, 2020.

\bibitem{Sarikaya2017TMI}
Duygu Sarikaya, Jason~J. Corso, and Khurshid~A. Guru,
\newblock ``{Detection and Localization of Robotic Tools in Robot-Assisted Surgery Videos Using Deep Neural Networks for Region Proposal and Detection},''
\newblock {\em T-MI}, 2017.

\bibitem{Bilen2016CVPR}
Hakan Bilen and Andrea Vedaldi,
\newblock ``{Weakly supervised deep detection networks},''
\newblock in {\em CVPR}, 2016.

\bibitem{Kantorov2016ECCV}
Vadim Kantorov, Maxime Oquab, Minsu Cho, and Ivan Laptev,
\newblock ``{Contextlocnet: Context-aware deep network models for weakly supervised localization},''
\newblock in {\em ECCV}, 2016.

\bibitem{Tang2018TPAMI}
Peng Tang, Xinggang Wang, Song Bai, Wei Shen, Xiang Bai, Wenyu Liu, and Alan Yuille,
\newblock ``{Pcl: Proposal cluster learning for weakly supervised object detection},''
\newblock {\em TPAMI}, 2018.

\bibitem{BearmanECCV2016}
Amy Bearman, Olga Russakovsky, Vittorio Ferrari, and Li~Fei-Fei,
\newblock ``{What's the Point: Semantic Segmentation with Point Supervision},''
\newblock in {\em ECCV}, 2016.

\bibitem{Jeong2019NEURIPS}
Jisoo Jeong, Seungeui Lee, Jeesoo Kim, and Nojun Kwak,
\newblock ``{Consistency-based Semi-supervised Learning for Object detection},''
\newblock in {\em NeurIPS}, H.~Wallach, H.~Larochelle, A.~Beygelzimer, F.~d\textquotesingle Alch\'{e}-Buc, E.~Fox, and R.~Garnett, Eds., 2019.

\bibitem{Sohn2020Arxiv}
Kihyuk Sohn, Zizhao Zhang, Chun-Liang Li, Han Zhang, Chen-Yu Lee, and Tomas Pfister,
\newblock ``{A simple semi-supervised learning framework for object detection},''
\newblock {\em ArXiv}, 2020.

\bibitem{Liu2021ICLR}
Yen-Cheng Liu, Chih-Yao Ma, Zijian He, Chia-Wen Kuo, Kan Chen, Peizhao Zhang, Bichen Wu, Zsolt Kira, and Peter Vajda,
\newblock ``{Unbiased Teacher for Semi-Supervised Object Detection},''
\newblock in {\em ICLR}, 2021.

\bibitem{Zhou2021CVPR}
Qiang Zhou, Chaohui Yu, Zhibin Wang, Qi~Qian, and Hao Li,
\newblock ``{Instant-teaching: An end-to-end semi-supervised object detection framework},''
\newblock in {\em CVPR}, 2021.

\bibitem{Vardazaryan2018MICCAI}
Armine Vardazaryan, Didier Mutter, Jacques Marescaux, and Nicolas Padoy,
\newblock ``{Weakly-supervised learning for tool localization in laparoscopic videos},''
\newblock in {\em MICCAI}, 2018.

\bibitem{Ali2022BBEngIV}
Mansoor Ali, Gilberto Ochoa-Ruiz, and Sharib Ali,
\newblock ``{A semi-supervised Teacher-Student framework for surgical tool detection and localization},''
\newblock {\em CMBBE}, 2022.

\bibitem{Yan2017Arxiv}
Ziang Yan, Jian Liang, Weishen Pan, Jin Li, and Changshui Zhang,
\newblock ``{Weakly- and Semi-Supervised Object Detection with Expectation-Maximization Algorithm},''
\newblock {\em ArXiv}, 2017.

\bibitem{Allan2020Robot}
Max Allan, Satoshi Kondo, Sebastian Bodenstedt, Stefan Leger, R~Kadkhodamohammadi, I~Luengo, Félix Fuentes, E~Flouty, A~Mohammed, M~Pedersen, Avinash Kori, V~Alex, G~Krishnamurthi, David Rauber, Robert Mendel, Christoph Palm, Sophia Bano, G~Saibro, C-S Shih, and Stefanie Speidel,
\newblock ``{2018 Robotic Scene Segmentation Challenge},''
\newblock 2020.

\bibitem{Ren2015NIPS}
Shaoqing Ren, Kaiming He, Ross Girshick, and Jian Sun,
\newblock ``{Faster R-CNN: Towards Real-Time Object Detection with Region Proposal Networks},''
\newblock in {\em NeurIPS}, 2015, vol.~28.

\bibitem{Lin2017CVPR}
Tsung-Yi Lin, Piotr Dollár, Ross Girshick, Kaiming He, Bharath Hariharan, and Serge Belongie,
\newblock ``Feature pyramid networks for object detection,''
\newblock in {\em CVPR}, 2017.

\bibitem{He2017ICCV}
Kaiming He, Georgia Gkioxari, Piotr Dollar, and Ross Girshick,
\newblock ``{Mask R-CNN},''
\newblock in {\em ICCV}, 2017.

\bibitem{Vaswani2017NIPS}
Ashish Vaswani, Noam Shazeer, Niki Parmar, Jakob Uszkoreit, Llion Jones, Aidan~N Gomez, \L~ukasz Kaiser, and Illia Polosukhin,
\newblock ``{Attention is All you Need},''
\newblock in {\em NeurIPS}, 2017.

\bibitem{bengio2013arxiv}
Samy Bengio, Jeff Dean, Dumitru Erhan, Eugene Ie, Quoc Le, Andrew Rabinovich, Jonathon Shlens, and Yoram Singer,
\newblock ``{Using web co-occurrence statistics for improving image categorization},''
\newblock {\em ArXiv}, 2013.

\bibitem{Lin2014ECCV}
Tsung-Yi Lin, Michael Maire, Serge Belongie, James Hays, Pietro Perona, Deva Ramanan, Piotr Doll{\'a}r, and C.~Lawrence Zitnick,
\newblock ``{Microsoft COCO: Common Objects in Context},''
\newblock in {\em ECCV}, 2014.

\bibitem{Gonzlez2020MICCAI}
Cristina Gonz{\'a}lez, Laura~Bravo S{\'a}nchez, and Pablo Arbel{\'a}ez,
\newblock ``{ISINet: An Instance-Based Approach for Surgical Instrument Segmentation},''
\newblock in {\em MICCAI}, 2020.

\bibitem{Sanchez2021MICCAI}
Ricardo Sanchez-Matilla, Maria Robu, Imanol Luengo, and Danail Stoyanov,
\newblock ``{Scalable Joint Detection and Segmentation of Surgical Instruments with Weak Supervision},''
\newblock in {\em MICCAI}, 2021.

\bibitem{Wu2019GitHub}
Yuxin Wu, Alexander Kirillov, Francisco Massa, Wan-Yen Lo, and Ross Girshick,
\newblock ``Detectron2,'' \url{https://github.com/facebookresearch/detectron2}, 2019.

\end{thebibliography}

\end{document}